\documentclass{article}

\usepackage[T1]{fontenc} 
\usepackage{blindtext}
\usepackage{graphicx}

\usepackage{indentfirst}

\title{Ontology for Healthcare Artificial Intelligence Privacy in Brazil}
\author{Vaz T.A., Dora J.M., Lamb L.C., Camey S.A.}
\date{December 2022}
\begin{document} 
\maketitle 
\begin{abstract}

\setlength{\lineskip}{3.5pt}
\setlength{\lineskiplimit}{2pt}
\setlength{\parindent}{26pt}
\setlength{\baselineskip}{12pt}

This article details the creation of a novel domain ontology at the intersection of epidemiology, medicine, statistics, and computer science. It outlines a systematic approach to handling structured data anonymously in preparation for its use in Artificial Intelligence (AI) applications in healthcare. The development followed 7 steps, including defining scope, selecting knowledge, reviewing important terms, constructing classes that describe designs used in epidemiological studies, machine learning paradigms, types of data and attributes, risks that anonymized data may be exposed to, privacy attacks, techniques to mitigate re-identification, privacy models, and metrics for measuring the effects of anonymization. It concludes with a practical implementation of this ontology in hospital settings to develop and validate AI systems.
   
\end{abstract}

\section{Introduction} 

The anonymity of Electronic Health Records (EHR) is critical for secondary use. Within the scope of AI research, anonymous data might still lack the necessary statistical properties to guarantee anonymity, posing challenges for researchers (SWEENEY, 2015, ROCHER, 2019). Proper treatment of this data is essential to maintain privacy and mitigate re-identification risks as required by the law (SPENGLER, 2019). In the world, nations have data protection laws that determine the mechanisms that need to be adopted processing personal data and sensitive personal data are conducted with special needs for information security, governance, and interoperability which in turn raises costs (SULLIVAN, 2004). However, some privacy laws state that anonymized data is no longer subject to these regulations (EU, 2016; BRAZIL, 2018).  

\begin{itemize}
    \item GDPR (Europe):  "information that does not relate to an identified or identifiable natural person or other personal data that has been anonymized." 
    
    \item LGPD (Brazil): "data relating to a holder who cannot be identified, considering the use of reasonable technical means available at the time of its treatment." 

\end{itemize}

To treat this discourse logically, it becomes necessary to semantically represent data from anonymized hospital records through a specific ontology (ABHYANKAR, 2012). This work presents the Ontology of Brazilian Hospital Records (ORHBR). It connects the LGPD with multidisciplinary concepts about privacy and data sciences from epidemiology, medicine, statistics, and computer sciences, forming the understanding of the necessary structures to describe the thinking about anonymizing hospital data.

\section{Methodology} To develop the ORHBR, we follow a 7-step methodology for constructing new ontologies. Step one defines the domain and scope, highlighting what is excluded. In step 2, knowledge selection occurs, including reviewing similar ontologies and determining the extension and adaptation proposal. In step 3, the essential terms of the scope and other standardizations in the health area are presented outside of this specific topic. In steps 4, 5, and 6, the work process for creating classes, properties, and relations occurs, defining them individually in a specialized tool for building ontologies. The last step is step 7, when we define an instance of the ontology (NOY, 2001).

\section{Development}

To develop the ORBHR following the proposed methodology, we performed the following steps using the protege.standford.edu tool (Figure 1) available on the internet. 

\begin{figure}
\centering
Figure 1 - ORHBR structure of classes and related properties 
\includegraphics[scale=0.26]{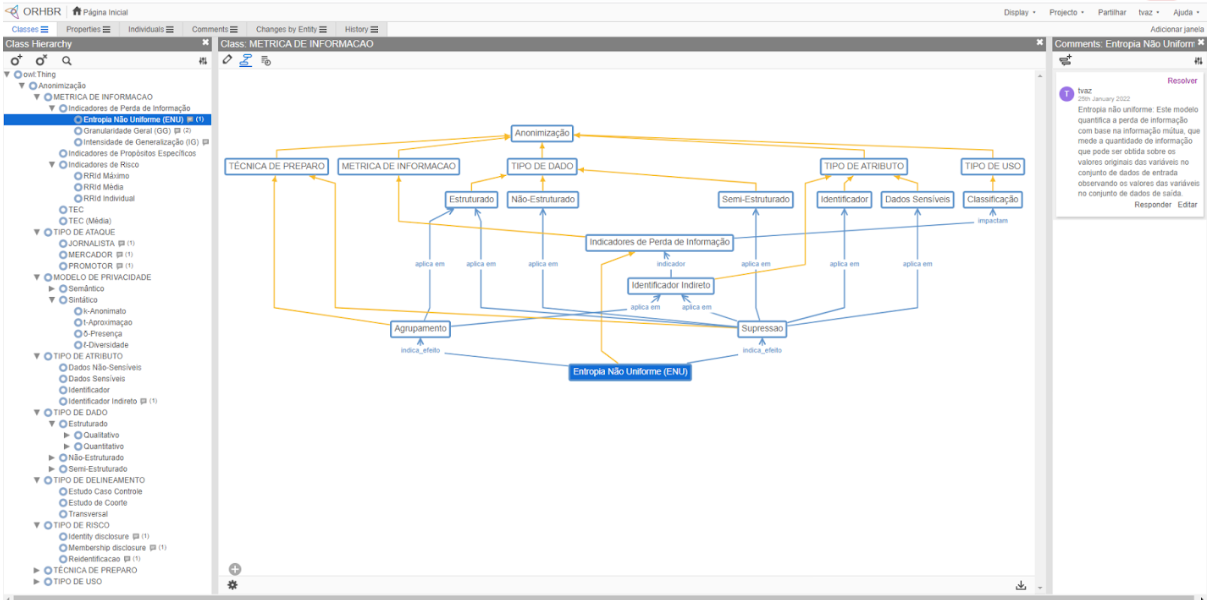}
Source: WebProtégé - Ontology development environment. Screenshot by Authors, 2022.
\end{figure}

\subsection{Definition of domain and scope}

The scope includes assessing whether anonymizing a hospital dataset is represented semantically appropriately.
Anonymization is defined as preparing a dataset to prevent re-identification of the data subject using reasonable technical means available during processing (BRAZIL, 2018). Other privacy methods, such as de-identification by removing predefined items (SULLIVAN, 2004), pseudonymization by replacing identifiers with an alternative key (OLIVEIRA, 2020), and cryptography (including hash functions and blockchain) that can be decrypted and identified (KRAWIEC, 2016), are outside the scope of this work. It intends to support researchers in answering qualitative questions to characterize different clinical and epidemiological observational studies that use anonymized hospital records.
The scope of patients' health is separate from the scope of this ontology, as it is not a new vocabulary of medical terms. ORHBR is also not an ontology for organizing information from EHR, it's a domain-specific ontology of anonymization methods for AI research, built as an agnostic to be used independently of other standards and conventions researchers may use.

\subsection{Knowledge Selection}

We use the work of QUEIROZ (2016) as a reference. It presents a domain ontology for preserving privacy in data published by the Brazilian federal government to control access to information and proposes the main classes necessary to address the topic. BATET (2011) established an ontology that defines metrics to compute semantic similarity in biomedicine, essential to define comparable results in this topic. LLUIS (2011) presents, in his doctoral thesis, an ontology for the statistical properties of anonymization within answering out data disturbance, proposing definitions of the types of possible data processing and the types of threats to which the data are exposed. A generic type of data called OntoDT established the concept of data typing (PANOV, 2016). The ORHBR proposal is born within the scope of the LGPD - General Data Protection Law (BRAZIL, 2018), while the reference selected for this work had its origin aiming at the protection of public documents of the federal government provided to the general public within the scope of the law Brazilian law that regulates access to information.

\subsection{Important Terms}

In Brazil, personal data refers to information related to an identified or identifiable natural person. Sensitive personal data includes genetic or biometric data, health information, racial or ethnic origin, and religious, philosophical, or political affiliations. Data processing includes all operations performed with personal data, such as collection, reception, classification, access, all forms of processing, storage in various media, and disposal (BRAZIL, 2018). Identifiers are data that directly identify the data subject. Indirect identifiers are data that can be cross-referenced with other public information sources. Standard terms for identifying data and indirect identifiers include: full name, address (subdivisions smaller than a state), dates related to an individual (birth, admission, discharge, death), contact information (phone numbers, e-mail addresses, social networks), identification codes (social security, health plan, medical record number, bank account, credit card, certificates, invoices, serial numbers of hospital products and medical devices), URL, cookies, IP number, username, biometric identifiers (fingerprint, retina, voice), images and sounds (including medical diagnostic images and facial photographs), and biological samples (SULLIVAN, 2004). Other important terms are derived from international standards and dictionaries: the dictionary of epidemiology by Prof. Miguel Porta, sponsored by the International Association of Epidemiology (PORTA, 2014); medical terms from the Unified Medical Language (UMLS); computer science terminology by KERR (2016); and the Health Informatics and Management Systems Society dictionary (HIMSS, 2013).

\section{Class Creation}
Classes represented the central structures of an ontology and were defined according to the needs identified during the requirements analysis to create the first instance. Next, we present the description of each class, conceptualizing the values that belong to each of their properties.
\subsection{Study Designs}

Within the scope of this ontology, we identified three types of designs used in epidemiological research studies that can be performed using anonymized secondary records. We will indicate references for details on the main epidemiological designs (NUNES; CAMEY; GUIMARÃES, 2013). 
\begin{itemize}
  \item Cross-sectional: an observational study examining data set at a given time to estimate the frequency of a given event. Exposure and outcome are collected from the database without information on event dates. They analyze the prevalence and association between exposure (exposed and unexposed) and the outcome class, which is usually binary but can present different data types. 

  \item Case-control: an observational study to compare people with a positive outcome of interest and a control group of people who do not have the outcome. It usually uses extended periods of data collection systematically. It allows for identifying risk factors in rare diseases, studying their etiology, and analyzing the odds ratio.

  \item Cohort: an observational study carried out to follow a group of people over time to assess the risks and benefits of using a given intervention or medication and to study the evolution and prognosis of diseases. Compares the disease incidence in the population using a ratio (relative risk) or a difference (attributable risk).
\end{itemize}

These types of studies have purposes and limitations that computer scientists need to address during experiments with AI applied to health. For example, the health data observed in an Electronic Health Record (EHR) do not record the patient's complete health events and require specific analyses to circumvent possible research biases. Furthermore, these data are subject to different types of systematic errors that reflect the quality of how hospital staff use the EHR. Other designs of epidemiological studies generally do not allow the systematic use of anonymization. Case reports describe isolated cases in detail and do not involve a data-centric analysis. Experimental studies (for example, randomized clinical trials) involve prospective follow-up of research subjects. Even if measures are taken to preserve privacy using data preparation, for security reasons related to the participants' health, most of these studies need ways of re-identifying the data subjects, making anonymization an unfeasible alternative.
 
\subsection{Data Type}

Within the scope of anonymization, we use the datasets to designate the structures containing rows and columns of data, also called tables, spreadsheets, or tabular data. We will call all columns variables, also called covariates or features. In order to understand the types of data in a hospital record, we classified the types of data into subclasses to represent different topics.

\begin{itemize}
\item Nature: Qualitative (ordinal and nominal), Quantitative (discrete and continuous). (YADAV, 2019)
\item Structure: Structured, semi-structured, unstructured, raw. (KIMBALL, 2011)
\item Computation: int, float, char (C++ language examples) (IBM, 1993).
\item Metadata: Taxonomy (labels) and dictionaries. (CHUTE, 2010)
\item Dataset Digital Format: Plaintext (UTF-8, ISO-9071 and others), Proprietary, Encrypted. (IBM, 1993)
\item Localization: Portuguese-BR, English-USA, and others define the language, time zone, and reference values that may vary depending on the location. (IBM, 1993)\end{itemize}
Having presented the data types that can be used, we can now classify a variable according to its perspective for anonymization, which we will call attribute types.

\subsection{Atribute Type}

The attributes define how each variable will be treated during the anonymization of the information.

\begin{itemize}
\item Identifiers: data that directly link the natural person who is the data subject. It includes but is not limited to the full name, patient's medical record, personal document number, and professional license record, which need to be removed during treatment for anonymization. (BRAZIL, 2018)
\item Indirect identifiers: when combined, can reveal the identity of personal data, but can be treated with a privacy algorithm to implement k-anonymity. (BRAZIL, 2018;)
\item Sensitive Personal Information: All other information about a person's health. They can be treated anonymously with the l-diversity and t-approximation algorithms. (MACHANAVAJJHALA, 2007)
\end{itemize}

From the definitions of the types of attributes, we can analyze the types of risk, the attacks, and the privacy models that can be used to mitigate the risks against privacy.

\subsection{Types of Risk}

The top three privacy-related risks that datasets are threatened with are:

\begin{itemize}

\item Identity disclosure (re-identification): this is a high privacy impact risk, as when an attacker successfully re-identifies, he learns all sensitive information about the data subject in the dataset. (SWEENEY, 2002)

\item Disclosure of attributes: this is a risk of intermediary impact on privacy because when an attacker is successful, only the value of some variables in the set is disclosed, which may allow inferring who the holders of the data contained in the set, but not pointing out who and who. (PRASSER, 2016)

\item Disclosure o association: this is a risk of lesser impact since the attacker does not directly disclose any information from the data set itself, but allows determining whether or not the holder is within the data set. (PRASSER, 2016)

\end{itemize}
\subsection{Attack Type}

Three attack models can be used to try to identify data that is considered anonymous.

\begin{itemize}
\item Journalist Model: attacks to disclose the identity of a specific data subject. It uses the data linkage technique to relate indirect identifiers in the dataset with other public information online (SWEENEY, 2015).

\item Prosecutor Model: attacks to disclose the identity of the data subject or a specific attribute, using as prior knowledge or background knowledge whether or not the data of interest are contained in the data set. (PRASSER, 2016)

\item Merchant Model: when there is no specific target but aims to identify many data subjects in a set. (PRASSER, 2016)

\end{itemize}

\subsection{Privacy Model}

An electronic hospital record can have re-identification risks mitigated with a privacy model: 

\begin{itemize}
\item k-Anonymity: uses grouping changes and suppression of indirect identifiers to ensure that an individual's data is indistinguishable from k-1 others (SWEENEY, 2002).

\item l-Diversity is an extension to k-anonymity insofar as it uses the same protection given to indirect identifiers and the set's sensitive personal data, increasing the computational complexity insofar as new categories need to be defined for each current value in the health datasets. (MACHANAVAJJHALA, 2007)

\item t-Approximation: proposes to overcome the limitation of l-diversity by redefining the distribution of the values of each variable instead of grouping. (RAJENDRAN, 2017)

\item O-Presence: can be used to protect data from membership disclosure, where a dataset reveals the probability that an individual from the population is contained in the dataset (RAJENDRAN, 2017).

\end{itemize}

Other models adapt the presented methods, which this work will not treat. (KHAN, 2021). With the privacy models defined, we can apply the related preparation techniques.

\subsection{Preparation Technique}

The decisive action to preserve the privacy of a data set is its preparation with one or more specific techniques used according to the selected privacy model:

\begin{itemize}
\item Suppression: Deletion of data that may indirectly identify a person. For this, algorithms are used that can entirely suppress or partially mask the observations, variables, or specific values within a data set (SAMARATI, 2012)

\item Grouping: Classification of patients into categories (also called generalization). The grouping of values allows all patients with the same category to be classified in the same groupings. (SAMARATI, 2012)

\item  Disturbance: Inclusion of noise or dirt, intentionally modifying data to make re-identification difficult. It is used to implement the differential privacy method in large databases. Due to the treatment that uses the modification of actual data, we will not deal with these models during this work (DWORK, 2008).

\end{itemize}

\subsection{Information Metric}

  By using the preparation techniques, we reduce the risk of re-identification and the loss of usefulness of the information. These metrics generally use original data (input) and anonymized data (output) to be computed.

We will use the following basic definitions:
\begin{itemize}

\item Indirect Identifiers: These are all variables the researcher defines as attributes that can be combined to re-identify a record. Usually, these attributes, such as age, gender, marital status, skin color, and level of education, can be found publicly.

\item Equivalence class: records with the same indirect identifiers have the same equivalence class (Ej). Therefore, D = E1 U ... U Ej, where j is the number of equivalence classes in the set D.
Class Equivalence Size (CES): is the number of records i that belong to the same equivalence class.

\item Hyperparameter k: the value k is defined a priori by the researcher and is used by the anonymization algorithms to limit the minimum size of the CES in an anonymized set. The greater the value of k, the greater the privacy of a set, while k=1 means that the records are not anonymized.

\end{itemize}
 
  For an original set D(\(A_{1}\),...,\(A_{p}\)) containing the indirect identifiers A = \(A_{1}\),...,\(A_ {p}\) suppressed or grouped into the anonymized set Dz(Az1,...,Azp) where CES(Dz, Az, i) >=k, for all i. Thus, we defined the metrics that measure the effect of anonymization.

\begin{itemize}
\item Individual Re-identification Risk (RR)
Is the individual re-identification risk of each record i in dataset D containing indirect identifiers A, depends on the CES value and is calculated using the formula:
  (SWEENEY, 2002)

\begin{equation}
     RR(D,A,i) = \frac{1}{CES(D,A,i)}*100
\end{equation}

\item Average Re-identification Risk(Avg RR)
It is the average re-identification risk of a set D containing n records i. The Average RR of the entire set is important, as the anonymization algorithms tend to preserve the Individual RR with values close to the Maximum RR, regardless of the techniques used in its preparation. However, the Average RR tends to be more susceptible to identifying improvements in the method of preparing indirect indicators, thus pointing out improvements in the risk of preserving the privacy of a set. Its value depends on the value of the Individual RR and is calculated with the formula:

\begin{equation}
      Average RR (D,A) = \frac{RR(D,A,i)+ ... +RR(D,A,n)}{n} *100
\end{equation}

\item Maximum Re-identification Risk
  (Maximum RR), which is the chance of success that an attack against privacy can have and re-identify at least one of the holders present in a data set, being:
 
   \begin{equation}
  Maximum RR(D) = \frac{1}{k}*100
  \end{equation}
 
\item Non-Uniform Entropy (NUE): compares the difference before and after the anonymization of the equivalence size of the classes in the whole dataset and individually for each attribute. The formula assumes the quotient is always less than or equal to 100, as the CES of i can only increase during preparation. Thus, the negative logarithm of the ratio is always a positive number, and the sum of all shows the attribute's entropy. The closer the value of NUE is to 1, the smaller the loss of variable information during the anonymization from D to Dz. (PRASSER; BILD; KUHN, 2016)

\begin{equation}
NUE(D,Ap,Dz, Azp) =(1 - (\sum_{i=1}^{n} - log \frac{CES (D, Ap, i)}{CES (D, Ap, i)} ) *100
\end{equation}

\item Intensity of Generalization (IG): identifies the loss of information between the original set and the anonymized set from the sum of the number of values modified during anonymization (SWEENEY, 2002). Where: I is the indicator function, that is, when AijAzij assumes value 1, otherwise it assumes value 0. n is the total number of lines, p is the total number of attributes, D is the original set and Dz is the anonymous set. Therefore, IG(D, Dz) = 1 if no indirect identifiers are modified. The more categorizations and deletions occur, the more the loss of information increases and the IG approaches zero. If all the values of the indirect indicators undergo suppression or categorization, we have IG(D, Dz) = 0. We use the IG to compute the number of values that were modified in the preparation, comparing the equality of the values before and after in all the rows i and columns j of both sets with the following formula:

\begin{equation}
     IG(D,Dz)=(1- \frac{\sum_{i=1}^{n} \sum_{j=1}^{n}I(Aij \neq Azij)}{n * p} ) * 100
\end{equation}

\item General Granularity (GG): compares the distinct amount of existing values in a variable before and after anonymization to show the loss of information. Where QAzn is the distinct number of values existing in Az after preparation, and QAn is the distinct number of values before preparation. The closer the value of GG is to 100, the smaller the loss of quality of information from A during anonymization. (IYENGAR, 2002)

\begin{equation}
     GG(Ap,Azp) = \frac{QAzp} {QAp} * 100
\end{equation}

\item Specific purpose method: compares models derived from different ways of preparing the same data set. For example, developing a classifier with machine learning and using the results of accuracy, sensitivity, and specificity, among others, to measure the effects of anonymization. (Pnumber, 2016)

\end{itemize}

\subsection{Type of Use }

Anonymizing a dataset impacts information loss indicators and, consequently, the results of using AI techniques. Therefore, the preparation needs to be done according to these indicators and considering the different types of use, as proposed by LEFEVRE (2008):

\begin{itemize}
\item Linear regression analysis: involves finding a linear model that describes or predicts the value of a quantitative dependent variable as a function of the other variables in the set. Linear regression analyses can be implemented with AI using supervised machine learning, including linear regression, neural networks, regression trees, and more. (JAMES, 2014)
\item Classification: is the attribution of qualitative variables that represent classes with predetermined values (also called targets, categories, dependent variables, targets, or labels) using a systematic procedure based on the observed variables. It is a task performed in cross-sectional studies and is characteristic of not using dates. AI classifier algorithms can use different learning approaches, including supervised, semi-supervised, and unsupervised. The main methods for classification include logistic regression, methods based on decision trees, neural networks, linear discriminant analysis, clustering, boosting, and support vector machines (SVM).
\item Information Retrieval: A selection (or query) involves a set of criteria used to filter data and define groups for a population (subpopulation). Combinations using logical operators allow the formulation of complex queries usually implemented with Structured Query Languages (SQL). The use of natural language processing (NLP), Neural Networks (NN), Large Language Models (LLM), and other AI techniques allows the selection of information in free text and images as well. Selection tasks are required during different times of a study. (LEFEVRE; DEWITT; RAMAKRISHNAN, 2008)
\item Clustering: involves recognizing, differentiating, and understanding how data can be grouped into categories. Clustering is also a type of staging used to implement anonymization, and for this reason, clustered data is also often considered anonymous data. However, some tasks derived from clustering that can be performed to analyze the specific demands for anonymization are topic classification, assignment of taxonomies, and clustering (which can be done using unsupervised machine learning) (IBM, 1993).
\end{itemize}

There are many ways to use an anonymized dataset. We focus on those who use Machine Learning (ML) to develop algorithms that can learn from their experiences and thus improve their performance in specific tasks necessary to solve problems. These tasks can be implemented with AI in different computer programs and via programming using libraries and machine learning packages Scikit Learn and Stats for Python and R, respectively. (KUHN, 2008; LORENZONI, 2019; PEDREGOSA, 2011; SANCHES, 2003)

\subsection{Creating properties}

With the definition of the classes and their subclasses, we can create the properties that indicate the existing relationships between the classes, answering questions using important terms for anonymizing hospital records, such as: What type of risk is the dataset exposed? What data type are the attributes that are indirect identifiers in the set? What privacy models mitigate the risk against the anticipated types of attacks? What are the risks of re-identifying the set in case of a specific kind of attack? What metrics can identify information loss in data prepared with suppression? Answering these questions, we can define properties as functions that connect classes, for example:

\begin{itemize}

\item has-preparation (Privacy Model, Preparation Technique): defines that a given privacy model can use certain data preparation techniques in its implementation, for example:

k-Anonymity <has-preparation> {Suppression, Grouping}

\item has-measure (Data Type, Information Metric): defines, for a data type, which metric shows the loss of information during anonymization:

Nominal Data Type <has-measure> NUE

\item has-impact (Information Metric, Task Type): defines which type of task performed by the AI can have the anonymization impact evaluated by a metric:

NUE <has-impact> Classification

\end{itemize}

Once the properties of the classes are defined, we can determine the relationships between them and an individual (or instance) in the real world.

\subsection{Creating relationships}

Individuals (or objects) in real life can use the terms of this ontology and thus create the necessary relationships to handle anonymization according to the purpose of use that the datasets will have (NOY, 2001). As an example, let us use a researcher who will study how to develop an AI application to predict hospital mortality, classifying which patients have the highest risk of dying in the first hours of hospitalization.

The first question to start this relationship between the researcher and the anonymization methods must be asked to understand the study design and thus identify its properties in the ontology. To start this conversation, what is the study design? From this answer, we use the properties that connect the classes and the properties in the anonymization domain to define a sequence of relationships to treat the subject in the context of a specific study. Thus, we can define an ontology instance.

\section{Results}

The instance creation is the seventh step, representing a real ontology application (NOY, 2001). We defined 5 instances of it in real medical settings, showing the following results:
\begin{itemize}
\item Instance A: 1) a cross-sectional study on hospital mortality, 2) the AI technique used to classify patients at risk of death, 3) the types of data used are structured, of a qualitative and quantitative nature, 4) qualitative data are strings and quantitative data are floats, 5) the medical record variable is a direct identifying attribute, 6) the age and sex attribute variables are indirect identifying attributes, 7) the objective is to mitigate the risk of re-identification, 8) defending of journalist-type attacks, 9) adopting the k-anonymity privacy model, 10) using suppression and grouping preparation techniques, 11) with the risk of re-identification measured by the metrics of RR=3.25, Average RR=2.57, and Maximum RR=6.81.

\item Instance B: 1) a cohort study on long-term medication effects, 2) the AI technique used to perform predictive modeling to assess risk factors and outcomes, 3) the types of data used are semi-structured 4) the types of data are character arrays, integers and floats, 5) the patient identification number is a direct identifying attribute, 6) the date of birth and geographic location attribute variables are indirect identifying attributes, 7) the objective is to ensure data privacy while analyzing long-term health trends, 8) defending of prosecutor-type attacks, 9) adopting the l-diversity privacy model, 10) using generalization and perturbation preparation techniques, 11) with the risk of re-identification measured by the metrics NUE=85.25, and IG=97.5.

\item Instance C: 1) a case-control study on rare disease etiology, 2) the AI technique used to perform machine learning to identify disease predictors, 3) the types of data used are unstructured, including medical notes and reports, 4) quantitative data are floats and qualitative data are strings, 5) the social security number is a direct identifying attribute, 6) the hospital admission and discharge dates attribute variables are indirect identifying attributes, 7) the objective is to prevent re-identification in sensitive disease research, 8) defending of merchant-type attacks, 9) adopting the t-closeness privacy model, 10) using noise addition and data masking preparation techniques, 11) with the risk of re-identification measured by Maximum RR=100, and GG=1.

\item Instance D: 1) a randomized clinical trial on new drug efficacy, 2) the AI technique used to perform statistical analysis to determine treatment effects, 3) the types of data used are structured and encrypted, both qualitative and quantitative, 4) qualitative data are strings and quantitative data are floats,  encoded values and encrypted floats, 5) the patient consent form is a direct identifying attribute, 6) the treatment start and end dates attribute variables are indirect identifying attributes, 7) the objective is to ensure patient privacy while maintaining data utility, 8) defending of prosecutor-type attacks, 9) adopting the O-Presence privacy model, 10) using data encryption and tokenization preparation techniques, 11) with the risk of re-identification measured by Specific Purpose Method metrics.

\item Instance E: 1) a prospective study on chronic disease management, 2) the AI technique used to perform time series analysis to monitor disease progression, 3) the types of data used are structured and semi-structured, with longitudinal entries, 4) quantitative data are floats, ordinal scales and of a quantitative nature are strings with time-stamped values, 5) the patient’s full name is a direct identifying attribute, 6) the chronic disease diagnosis dates and treatment intervals attribute variables are indirect identifying attributes, 7) the objective is to protect patient identities over extended study periods, 8) defending of journalist-type attacks, 9) adopting the k-anonymity privacy model with l-diversity enhancement, 10) using data aggregation and differential privacy preparation techniques, 11) with the risk of re-identification measured by Specific Purpose Method metrics.

\end{itemize}

\section{Discussion}

In their review of the topic, CHEVRIER (2019) deals with the variability observed in the terms that define the methods of preparation and in the way in which the terms de-identification and anonymization are used, emphasizing the need for objective definitions centered on the legislation to improve education and dissemination of information on the subject (CHEVRIER, 2019). This ontology defines a formal structure to understand the risk of re-identification of hospital records, often translating different concepts between different areas, and this required making choices. Our expectation is not to change the culture of consolidated research areas but to help form a new community of researchers who will use hospital datasets not only to train and use AI but to enhance the use of data in different types of epidemiological studies, hospital management and innovation in health as a whole. There are legal provisions for using hospital data; one is the patient's informed consent, authorizing the use of their data in research (BRAZIL, 2018). To obtain informed consent from patients, it is necessary to interact with them. How this consent is obtained can introduce more than one type of bias in clinical research (EMAM, 2013). Anonymizing datasets at the first opportunity after collection is an alternative to not needing to obtain this consent. However, data processing for anonymization can interfere with the usefulness of the data and prevent the objective research search from being achieved. Thus, ORHBR also describes the semantics to explain why a given research needs to work with identified data. Our study could have been more extensive in elaborating on only 5 instance used to exemplify the ontology. We understand that this is the beginning of a new field in health research, and creating other types of instances may demand new requirements. We believe that an ontology such as ORHBR can be adopted by the national authority for the protection of personal data and deposited in the Electronic Government Vocabulary and Ontology Repository, enhancing the use by other researchers and maintenance of the ontology, aiming to extend its classes by including new types of data (for example image and sound) and types of design (for instance counterfactual studies), supporting the maintenance of an anonymization policy that enhances the use of data.

\section{Conclusion}

We defined an ontology to leverage the culture of privacy in research with hospital records while maintaining sight of the opportunities to solve problems of different types using AI. By adopting ORHBR, researchers from other areas can share and reuse technical knowledge about anonymizing hospital records using the same vocabulary. Using this ontology, we can quantitatively and qualitatively compare different privacy models that inform the risks and loss of information during the anonymization process.

\end{document}